\documentclass[11pt,a4paper]{article}
\usepackage[hyperref]{emnlp2020}
\usepackage{times}
\usepackage{latexsym}

\usepackage{microtype}
\usepackage{booktabs}
\usepackage{graphicx}
\usepackage{amsmath, amssymb, amsfonts}
\usepackage{float}

\usepackage[]{todonotes}

\makeatother

\aclfinalcopy %
\def\aclpaperid{3373} %

\title{An Empirical Investigation of Beam-Aware Training in Supertagging}

\author{Renato Negrinho$^1$ \\
  \\
  \\\And
  Matthew R. Gormley$^1$ \\
  Carnegie Mellon University$^1$, MSR Montreal$^2$ \\
  \texttt{\{negrinho,mgormley,ggordon\}@cs.cmu.edu}
  \\\And
  Geoffrey J. Gordon$^{1, 2}$ \\
  \\
  \\
   }

\date{}

\begin{document}
\maketitle

\begin{abstract}
  Structured prediction is often approached by training a locally normalized model with maximum likelihood and decoding approximately with beam search.
  This approach leads to mismatches as, during training, the model is not exposed to its mistakes and does not use beam search.
  Beam-aware training aims to address these problems, but unfortunately, it is not yet widely used due to a lack of understanding about how it impacts performance, when it is most useful, and whether it is stable.
  Recently, \citet{negrinho2018learning} proposed a meta-algorithm that captures beam-aware training algorithms and suggests new ones, but unfortunately did not provide empirical results.
  In this paper, we begin an empirical investigation: we train the supertagging model of \citet{vaswani2016supertagging} and a simpler model with instantiations of the meta-algorithm.
  We explore the influence of various design choices and make recommendations for choosing them.
  We observe that beam-aware training improves performance for both models, with large improvements for the simpler model which must effectively manage uncertainty during decoding.
  Our results suggest that a model must be learned with search to maximize its effectiveness.
\end{abstract}

\section{Introduction}
\label{sec:introduction}

Structured prediction often relies on models that train on maximum likelihood and use beam search for approximate decoding.
This procedure leads to two significant mismatches between the training and testing settings:
the model is trained on oracle trajectories and therefore does not learn about its own mistakes;
the model is trained without beam search and therefore does not learn how to use the beam effectively for search.

Previous algorithms have addressed one or the other of these mismatches.
For example, DAgger~\cite{ross2011reduction} and scheduled sampling~\cite{bengio2015scheduled} use the learned model to visit non-oracle states at training time, but do not use beam search (i.e., they keep a single hypothesis).
Early update~\cite{collins_incremental_2004}, LaSO~\cite{daume2005learning}, and BSO~\cite{wiseman2016sequence} are trained with beam search, but do not expose the model to beams without a gold hypothesis (i.e., they either stop or reset to beams with a gold hypothesis).

Recently, \citet{negrinho2018learning} proposed a meta-algorithm that instantiates beam-aware algorithms as a result of choices for the surrogate loss (i.e., which training loss to incur at each visited beam) and data collection strategy (i.e., which beams to visit during training).
A specific instantiation of their meta-algorithm addresses both mismatches by relying on an insight on how to induce training losses for beams without the gold hypothesis: \emph{for any beam}, its lowest cost neighbor should be scored sufficiently high to be kept in the successor beam.
To induce these training losses it is sufficient to be able to compute the best neighbor of any state (often called a dynamic oracle~\cite{goldberg2012dynamic}).
Unfortunately, \citet{negrinho2018learning} do not provide empirical results, leaving open questions such as whether instances can be trained robustly, when is beam-aware training most useful, and what is the impact on performance of the design choices.

\paragraph{Contributions}
We empirically study beam-aware algorithms instantiated through the meta-algorithm of~\citet{negrinho2018learning}.
We tackle supertagging as it is a sequence labelling task with an easy-to-compute dynamic oracle and a moderately-sized label set (approximately $1000$) which may require more effective search.
We examine two supertagging models (one from \citet{vaswani2016supertagging} and a simplified version designed to be heavily reliant on search) and train them with instantiations of the meta-algorithm.
We explore how design choices influence performance, and give recommendations based on our empirical findings.
For example, we find that perceptron losses perform consistently worse than margin and log losses.
We observe that beam-aware training can have a large impact on performance, particularly when the model must use the beam to manage uncertainty during prediction.
Code for reproducing all results in this paper is available at \url{https://github.com/negrinho/beam_learn_supertagging}.

\begin{figure*}[tbh]
  \centering
  \includegraphics[width=0.75\textwidth]{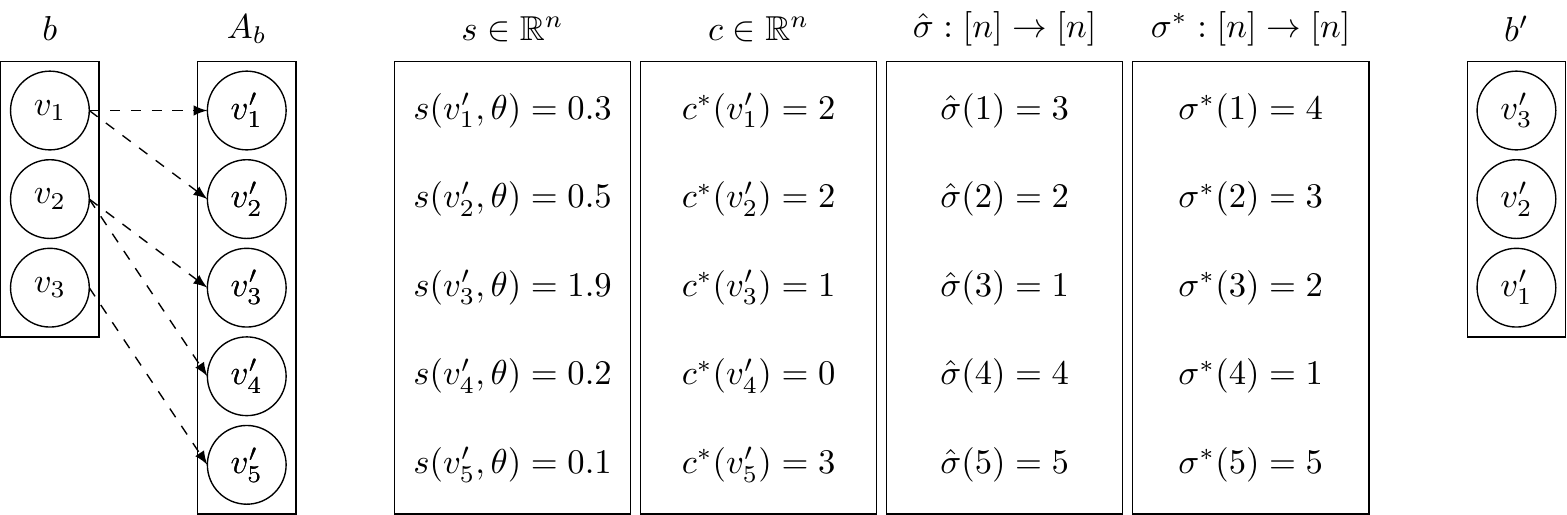}
  \caption{Beam $b$ has neighborhood $A_b$, where $k = |b| = |b'| = 3$ and $n = |A_b| = 5$.
  Edges from elements in $b$ to elements in $A_b$ encode neighborhood relationships, e.g., $v_3$ has a single neighbor $v'_5$.
  Permutation $\hat \sigma : [n] \to [n]$ sorts hypotheses in decreasing order of score, and permutation $\sigma^* : [n] \to [n]$ sorts them in increasing order of cost, i.e, $v'_{\sigma^*(1)}$ is the lowest cost neighbor and $v'_{\hat \sigma(1)}$ is the highest scoring neighbor.
  The successor beam $b'$ keeps the neighbor states in $A_b$ with highest score according to vector $s$, or equivalently highest rank according to $\hat \sigma$. %
  }
  \label{fig:local_notation}
\end{figure*}

\begin{figure*}[tbh]
    \centering
    \includegraphics[width=0.68\textwidth]{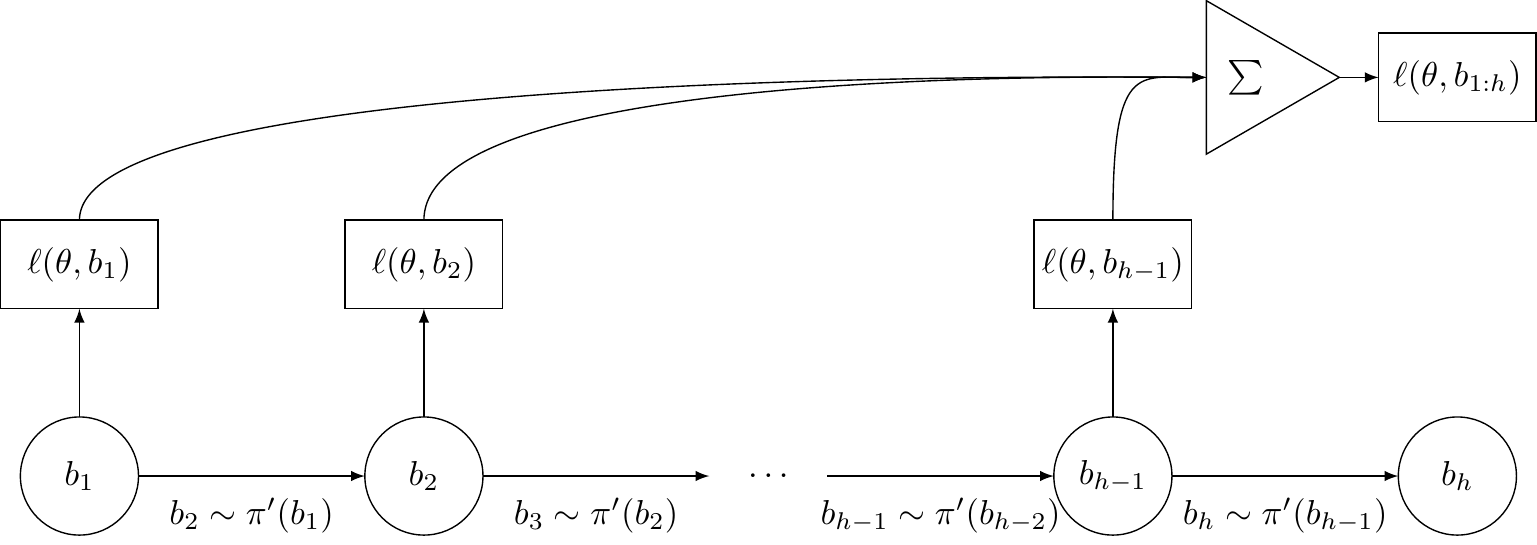}
    \caption{Sampling a trajectory through the beam search space at training time.
    A loss $\ell(b_i, \theta)$ is incurred at each visited beam $b_i$, $i \in [h - 1]$, resulting in total accumulated loss $\ell(b_{1:h}, \theta)$ for beam trajectory $b_{1:h}$.
    The terminal beam $b_h$ corresponds to a complete output $y(b_h) \in \mathcal Y$.
    Transitions between beams are sampled according to a data collection policy $\pi' : V_k \to \Delta(V_k)$.
    We consider $\pi'$ induced by a scoring function $s(\cdot, \theta) : V \to \mathbb R$ or cost function $c^* : V \to \mathbb R$.
    Parameters $\theta$ parametrize the scoring function of the model.
    Losses $\ell(b_i, \theta)$ are low if the scores of the neighbors of $b_i$ comfortably keep the lowest cost elements in the successor beam (see Section~\ref{ssec:surrogate_losses}), and high otherwise.
    See Figure~\ref{fig:local_notation} for the notation to describe the surrogate loss $\ell(b_i, \theta)$ at each beam $b_i$.
    }
    \label{fig:trajectory}
\end{figure*}

\section{Background on learning to search and beam-aware methods}
\label{sec:background}

For convenience, we reuse notation introduced in~\citet{negrinho2018learning} to describe their meta-algorithm and its components (e.g., scoring function, surrogate loss, and data collection strategy).
See Figure~\ref{fig:local_notation} and Figure~\ref{fig:trajectory} for an overview of the notation.
When relevant, we instantiate notation for left-to-right sequence labelling under the Hamming cost, which supertagging is a special case of.

\paragraph{Input and output spaces}
Given an input structure $x \in \mathcal X$, the output structure $y \in \mathcal Y_x$, is generated through a sequence of incremental decisions.
An example $x \in \mathcal X$ induces a tree $G_x = (V_x, E_x)$ encoding the sequential generation of elements in $\mathcal Y_x$, where $V_x$ is the set of nodes and $E_x$ is the set of edges.
The leaves of $G_x$ correspond to elements of $\mathcal Y_x$ and the internal nodes correspond to incomplete outputs.
For left-to-right sequence labelling, for a sequence $x \in \mathcal X$, each decision assigns a label to the current position of $x$ and the nodes of tree encode labelled prefixes of $x$, with terminal nodes encoding complete labellings of $x$.

\paragraph{Cost functions}
Given a golden pair $(x, y) \in \mathcal X \times \mathcal Y$, the cost function $c _{x, y} : \mathcal Y_x \to \mathbb R$ measures how bad the prediction $\hat y \in \mathcal Y_x$ is relative to the target output structure $y \in \mathcal Y_x$.
Using $c _{x, y} : \mathcal Y_x \to \mathbb R$, we define a cost function $c^*_{x,y} : V _x \to \mathbb R$ for partial outputs by assigning to each node $v \in V_x$ the cost of its best reachable complete output, i.e., $c _{x, y} ^*(v) = \min _{v' \in T_v} c _{x, y}(v')$, where $T_v \subseteq \mathcal Y_x$ is the set of complete outputs reachable from $v$.
For a left-to-right search space for sequence labelling, if $c_{x, y} : \mathcal Y_x \to \mathbb R$ is Hamming cost, the optimal completion cost $c^*_{x, y} : \mathcal Y_x \to \mathbb R$ is the number of mistakes in the prefix as the optimal completion matches the remaining suffix of the target output.

\paragraph{Dynamic oracles}
An oracle state is one for which the target output structure can be reached.
Often optimal actions can only be computed for oracle states.
Dynamic oracles compute optimal actions even for non-oracle states.
Evaluations of $c^* _{x, y} : V _x \to \mathbb R$ for arbitrary states allows us to induce the dynamic oracle---at a state $v \in V_x$, the optimal action is to transition to the neighbor $v' \in N_v$ with the lowest completion cost.
For sequence labelling, this picks the transition that assigns the correct label.
For other tasks and metrics, more complex dynamic oracles may exist, e.g., in dependency parsing~\citep{goldberg2012dynamic,goldberg2013training}.
For notational brevity, from now on, we omit the dependency of the search spaces and cost function on $x \in \mathcal X$, $y \in \mathcal Y$, or both.

\paragraph{Beam search space}
Given a search space $G = (V, E)$, the beam search space $G _k  = (V_k, E_k)$ is induced by choosing a beam size $k \in \mathbb N$ and a strategy for generating the successor beam out of the current beam and its neighbors.
In this paper, we expand all the elements in the beam and score the neighbors simultaneously.
The highest scoring $k$ neighbors are used to form the successor beam.
For $k = 1$, we recover the greedy search space $G$.

\paragraph{Beam cost functions}
The natural cost function $c^* : V_k \to \mathbb R$ for $G _k$ is created from the element-wise cost function on $G$, and assigns to each beam the cost of its best element, i.e., for $b \in V_k$, $c^*(b) = \min_{v \in b} c^*(v)$.
For a transition $(b, b') \in E_k$, we define the transition cost $c(b, b') = c^*(b') - c^*(b)$, where $b' \in N_b$, i.e., $b'$ can be formed from the neighbors of the elements in $b$.
A cost increase happens when $c(b, b') > 0$, i.e., the best complete output reachable in $b$ is no longer reachable in $b'$.

\paragraph{Policies}
Policies operate in beam search space $G _k$ and are induced through a learned scoring function $s(\cdot, \theta) : V \to \mathbb R$ which scores elements in the original space $G$.
A policy $\pi: V_k \to \Delta(V_k)$, i.e., mapping states (i.e., beams) to distributions over next states.
We only use deterministic policies where the successor beam is computed by sorting the neighbors in decreasing order of score and taking the top $k$.

\paragraph{Scoring function}
In the non-beam-aware case, the scoring function arises from the way probabilities of complete sequences are computed with the locally normalized model, namely
  $
  p(y|x, \theta) = \prod _{j = 1}^h p(y_i | y_{1:i-1}, x, \theta)
  $,
where we assume that all outputs for $x \in \mathcal X$ have $h$ steps.
For sequence labelling, $h$ is the length of the sentence.
The resulting scoring function $s(\cdot, \theta) : V \to \mathbb R$ is
  $
  s(v, \theta) = \sum _{i = 1}^j \log p(y_i | y_{1:i-1}, x, \theta)
  $,
where $v = y_{1:j}$ and $j \in [h]$.
Similarly, the scoring function that we learn in the beam-aware case is
  $
  s(v, \theta) = \sum _{i = 1}^j \tilde s (v, \theta)
  $,
where $x$ has been omitted, $v = y_{1:j}$, and $\tilde s(\cdot, \theta) : V \to \mathbb R$ is the learned incremental scoring function.
In Section~\ref{sec:discussion}, we observe that this cumulative version performs uniformly better than the non-cumulative one.

\section{Meta-algorithm for learning beam search policies}
\label{sec:meta_algorithm}

We refer the reader to \citet{negrinho2018learning} for a discussion of how specific choices for the meta-algorithm recover algorithms from the literature.

\subsection{Data collection strategies}
\label{ssec:data_collection}

The data collection strategy determines which beams are visited at training time (see Figure~\ref{fig:trajectory}).
Strategies that use the learned model differ on how they compute the successor beam $b' \in N_b$ when $s(\cdot, \theta)$ leads to a beam without the gold hypothesis, i.e., $c(b, b') > 0$, where $b' = \{v_{\hat \sigma(1)}, \ldots, v_{\hat \sigma (k)} \} \subset A_b$ and $A_b = \{v_1, \ldots, v_n\} = \cup_{v \in b} N_v$.
We explore several data collection strategies:

\paragraph{stop}
If the successor beam does not contain the gold hypothesis, stop collecting the trajectory.
Structured perceptron training with early update~\cite{collins_incremental_2004} use this strategy.

\paragraph{reset}
If the successor beam does not contain the gold hypothesis, reset to a beam with only the gold hypothesis\footnote{Any beam with the gold hypothesis would be valid, e.g., the top $k - 1$ according to the scores plus the gold hypothesis, which we call \textit{reset (multiple)}}.
LaSO~\cite{daume2005learning} use this strategy.
For $k = 1$, we recover teacher forcing as only the oracle hypothesis is kept in the beam.

\paragraph{continue}
Ignore cost increases, always using the successor beam.
DAgger~\cite{ross2011reduction} take this strategy, but does not use beam search.
\citet{negrinho2018learning} suggest this for beam-aware training but do not provide empirical results.

\paragraph{reset (multiple)}
Similar to reset, but keep $k - 1$ hypothesis from the transition, i.e., $b' = \{v_{\sigma^*(1)}, v_{\hat \sigma(1)}, \ldots v_{\hat \sigma(k - 1)}\}$.
We might expect this data collection strategy to be closer to \textit{continue} as a large fraction of the elements of the successor beam are induced by the learned model.

\paragraph{oracle}
Always transition to the beam induced by $\sigma^*: [n] \to [n]$, i.e., the one obtained by sorting the costs in increasing order.
For $k = 1$, this recovers teacher forcing.
In Section~\ref{ssec:comp_data_collection}, we observe that \textit{oracle} dramatically degrades performance due to increased exposure bias with increased $k$.

\subsection{Surrogate losses}
\label{ssec:surrogate_losses}

Surrogate losses encode that the scores produced by the model for the neighbors must score the best neighbor sufficiently high for it to be kept comfortably in the successor beam.
For $k = 1$, many of these losses reduce to losses used in non-beam-aware training.
Given scores $s \in \mathbb R^n$ and costs $c \in \mathbb R^n$ over neighbors in $A_b = \{v_1, \ldots, v_n\}$, we define permutations $\hat \sigma: [n] \to [n]$ and $\sigma^*: [n] \to [n]$ that sort the elements in $A_b$ in decreasing order of scores and increasing order of costs, respectively, i.e., $s_{\hat \sigma(1)} \geq \ldots \geq s_{\hat \sigma(n)}$ and $c_{\sigma^*(1)} \leq \ldots \leq s_{\sigma^*(n)}$.
See Figure~\ref{fig:local_notation} for a description of the notation used to describe surrogate losses.
Our experiments compare the following surrogate losses:

\paragraph{perceptron (first)} Penalize failing to score the best neighbor at the top of the beam (regardless of it falling out of the beam or not).
\begin{align*}
\ell(s, c) =
        \max
          \left( 0, s _{ \hat \sigma (1) } - s _{ \sigma ^*(1) } \right).
\end{align*}

\paragraph{perceptron (last)} If this loss is positive at a beam, the successor beam induced by the scores does not contain the golden hypothesis.
\begin{align*}
\ell(s, c) = \max
    \left(
        0, s _{ \hat \sigma (k) } - s _{ \sigma ^*(1) }
    \right).
\end{align*}

\paragraph{margin (last)} Penalize margin violations of the best neighbor of the hypothesis in the current beam.
Compares the correct neighbor $s_{\sigma^*(1)}$ with the neighbor $v _{ \hat \sigma(k)}$ last in the beam.
\begin{align*}
\ell(s, c) = \max
    \left(
        0, s _{ \hat \sigma(k) } - s _{ \sigma ^*(1) } + 1
    \right)
\end{align*}

\paragraph{cost-sensitive margin (last)}
Same as \textit{margin (last)} but weighted by the cost difference of the pair.
\citet{wiseman2016sequence} use this loss.
\begin{align*}
\ell(s, c) =
    (c _{\hat \sigma (k) } - c _{ \sigma ^* (1) })
      \max
        ( 0, s _{ \hat \sigma (k) } - s _{ \sigma ^*(1) } +1 ).
\end{align*}

\paragraph{log loss (neighbors)}
    Normalizes over all elements in $A_b$. For beam size $k = 1$, it reduces to the usual log loss.
    \begin{align*}
          \ell(s, c)
          &= - s _{ \sigma ^*(1) } + \log
          \left(
              { \sum _{ i = 1} ^n \exp ( s _i ) }
          \right)
  \end{align*}

\paragraph{log loss (beam)}
Normalizes only over the top $k$ neighbors of a beam according to the scores $s$.
  \begin{align*}
        \ell(s, c)
          &= - s _{ \sigma ^*(1) } + \log
          \left(
              \sum _{ i \in I }  \exp ( s _i )
          \right) ,
  \end{align*}
 where $I =\{ \sigma ^*(1), \hat \sigma(1), \ldots, \hat \sigma(k) \}$.
 The normalization is only over the golden hypothesis $v _{\sigma ^*(1)}$ and the elements included in the beam.
 \citet{andor2016globally} use this loss.

\begin{figure*}[tbp]
  \centering
  \includegraphics[width=0.8\textwidth]{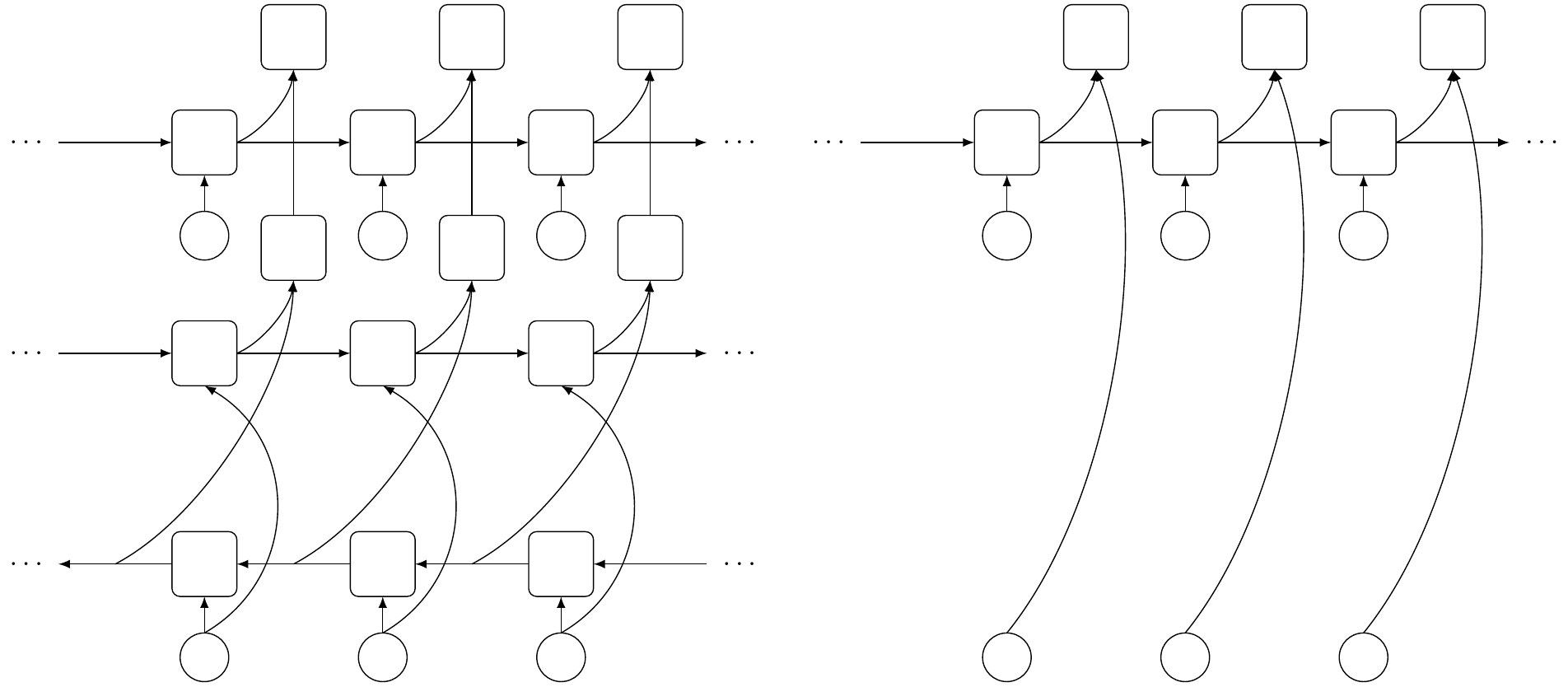}
  \caption{
    High-level structure of the two models used in the experiments.
    The model on the left is from \citet{vaswani2016supertagging}.
    The model on the right is a simplification of the one on the left, namely, it does not have an encoding of the complete sentence at the start of prediction.
  }
  \label{fig:model}
\end{figure*}

\subsection{Training}
\label{ssec:training}

The meta-algorithm of \citet{negrinho2018learning} is instantiated by choosing a surrogate loss, data collection strategy, and beam size.
Training proceeds by sampling an example $(x, y) \in \mathcal X \times \mathcal Y$ from the training set.
A trajectory through the beam search space $G_k$ is collected using the chosen data collection strategy.
A surrogate loss is induced at each non-terminal beam in the trajectory (see Figure~\ref{fig:trajectory}).
Parameter updates are computed based on the gradient of the sum of the losses of the visited beams.

\section{Experiments}
\label{sec:experiments}

We explore different configurations of the design choices of the meta-algorithm to understand their impact on training behavior and performance.

\subsection{Task details}
\label{ssec:task_details}

We train our models for supertagging, a sequence labelling where accuracy is the performance metric of interest.
Supertagging is a good task for exploring beam-aware training, as contrary to other sequence labelling datasets such as named-entity recognition~\citep{tjong_kim_sang_introduction_2003}, chunking~\citep{sang2000introduction}, and part-of-speech tagging~\citep{marcus_building_1993}, has a moderate number of labels and therefore it is likely to require effective search to achieve high performances.
We used the standard splits for CCGBank~\citep{hockenmaier2007ccgbank}: the training and development sets have, respectively, $39604$ and $1913$ examples.
Models were trained on the training set and used the development set to compute validation accuracy at the end of each epoch to keep the best model.
As we are performing an empirical study, similarly to \citet{vaswani2016supertagging}, we report validation accuracies.
Each configuration is ran three times with different random seeds and the mean and standard deviation are reported.
We replace the words that appear at most once in the training set by \texttt{UNK}.
By contrast, no tokenization was done for the training supertags.

\subsection{Model details}
\label{ssec:model_details}

We have implemented the model of~\citet{vaswani2016supertagging} and a simpler model designed by removing some of its components.
The two main differences between our implementation and theirs are that we do not use pretrained embeddings (we train the embeddings from scratch) and we use the gold POS tags (they use only the pretrained embeddings).
\paragraph{Main model}
For the model of \citet{vaswani2016supertagging} (see Figure~\ref{fig:model}, left), we use $64$, $16$, and $64$ for the dimensions of the word, part-of-speech, and supertag embeddings, respectively.
All LSTMs (forward, backward, and LM) have hidden dimension $256$.
We refer the reader to \citet{vaswani2016supertagging} for the exact description of the model.
Briefly, embeddings for the words and part-of-speech tags are concatenated and fed to a bi-directional LSTM, the outputs of both directions are then fed into a combiner (dimension-preserving linear transformations applied independently to both inputs, added together, and passed through a ReLU non-linearity).
The output of the combiner and the output of the LM LSTM (which tracks the supertag prefix up to a prediction point) is then passed to another combiner that generates scores over supertags.

\paragraph{Simplified model}
We also consider a simplified model that drops the bi-LSTM encoder and the corresponding combiner (see Figure~\ref{fig:model}, right).
The concatenated embeddings are fed directly into the second combiner with the LM LSTM output.
Values for the hyperparameters are the same when possible.
This model must leverage the beam effectively as it does not encode the sentence with a bi-LSTM.
Instead, only the embeddings for the current position are available, giving a larger role to the LM LSTM over supertags.
While supertagging can be tackled with a stronger model, this restriction is relevant for real-time tasks, e.g., the complete input might not be known upfront.

\paragraph{Training details}
Models are trained for $16$ epochs with SGD with batch size $1$ and cosine learning rate schedule~\citep{loshchilov2016sgdr}, starting at $10^{-1}$ and ending at $10^{-5}$.
No weight decay or dropout was used.
Training examples are shuffled after each epoch.
Results are reported for the model with the best validation performance across all epochs.
We use $16$ epochs for all models for simplicity and fairness.
This number was sufficient, e.g.,
we replicated Table~\ref{tab:results_strategies} by training with $32$ epochs and observed minor performance differences (see Table~\ref{tab:results_strategies_train32epochs}).

\subsection{Non-beam-aware training}
\label{ssec:nonbeam_aware_training}

We first train the models with $k = 1$ and then use beam search to decode.
Crucially, the model does not train with a beam and therefore does not learn to use it effectively.
We vary the data collection strategy.
The results are presented in Table~\ref{tab:results_nonbeam_aware} and should be used as a reference when reading the other tables to evaluate the impact of beam-aware training.
Tables are formatted such that the first and second horizontal halves contain the results for the main model and simplified model, respectively.
Each position contains the mean and the standard deviation of running that configuration three times.
We use this format in all tables presented.
%


\begin{table}[tb]
	\centering
  \resizebox{\columnwidth}{!}{
	\begin{tabular}{lcccc}
		\toprule
	    & 1 & 2 & 4 & 8 \\
		\midrule
      oracle/reset &  $93.78_{0.12}$ & $93.81_{0.11}$ & $93.82_{0.10}$ & $93.82_{0.10}$ \\
      continue &      $94.04_{0.07}$ & $94.05_{0.07}$ & $94.05_{0.07}$ & $94.06_{0.07}$ \\
      stop &          $93.86_{0.09}$ & $93.90_{0.07}$ & $93.90_{0.07}$ & $93.91_{0.07}$ \\
    \midrule
      oracle/reset &  $73.20_{0.31}$ & $76.55_{0.24}$ & $77.42_{0.27}$ & $77.54_{0.22}$ \\
		  continue &      $81.99_{0.04}$ & $82.30_{0.03}$ & $82.37_{0.08}$ & $82.41_{0.08}$ \\
      stop &          $74.35_{0.23}$ & $77.06_{0.14}$ & $77.73_{0.13}$ & $77.82_{0.09}$ \\
    \bottomrule
  \end{tabular}
  }
	\caption{
		Development accuracies for models trained with different data collection strategies in a non-beam-aware way (i.e., $k = 1$) and decoded with beam search with varying beam size.
    \textit{continue} performs best, showing the importance of exposing the model to its mistakes.
    Differences are larger for the simplified model.
    }
	\label{tab:results_nonbeam_aware}
\end{table}

The \textit{continue} data collection strategy (i.e., DAgger for $k = 1$) results in better models than training on the oracle trajectories.
Beam search results in small gains for these settings.
In this experiment, training with oracle is the same as training with reset as the beam always contains only the oracle hypothesis.
The performance differences are small for the main model but much larger for the simplified model, underscoring the importance of beam search when there is greater uncertainty about predictions.
For the stronger model, the encoding of the left and right contexts with the bi-LSTM provides enough information at each position to predict greedily, i.e., without search.
This difference appears consistently in all experiments, with larger gains for the weaker model.

\begin{figure*}[tbp]
  \centering
  \includegraphics[width=0.24\textwidth]{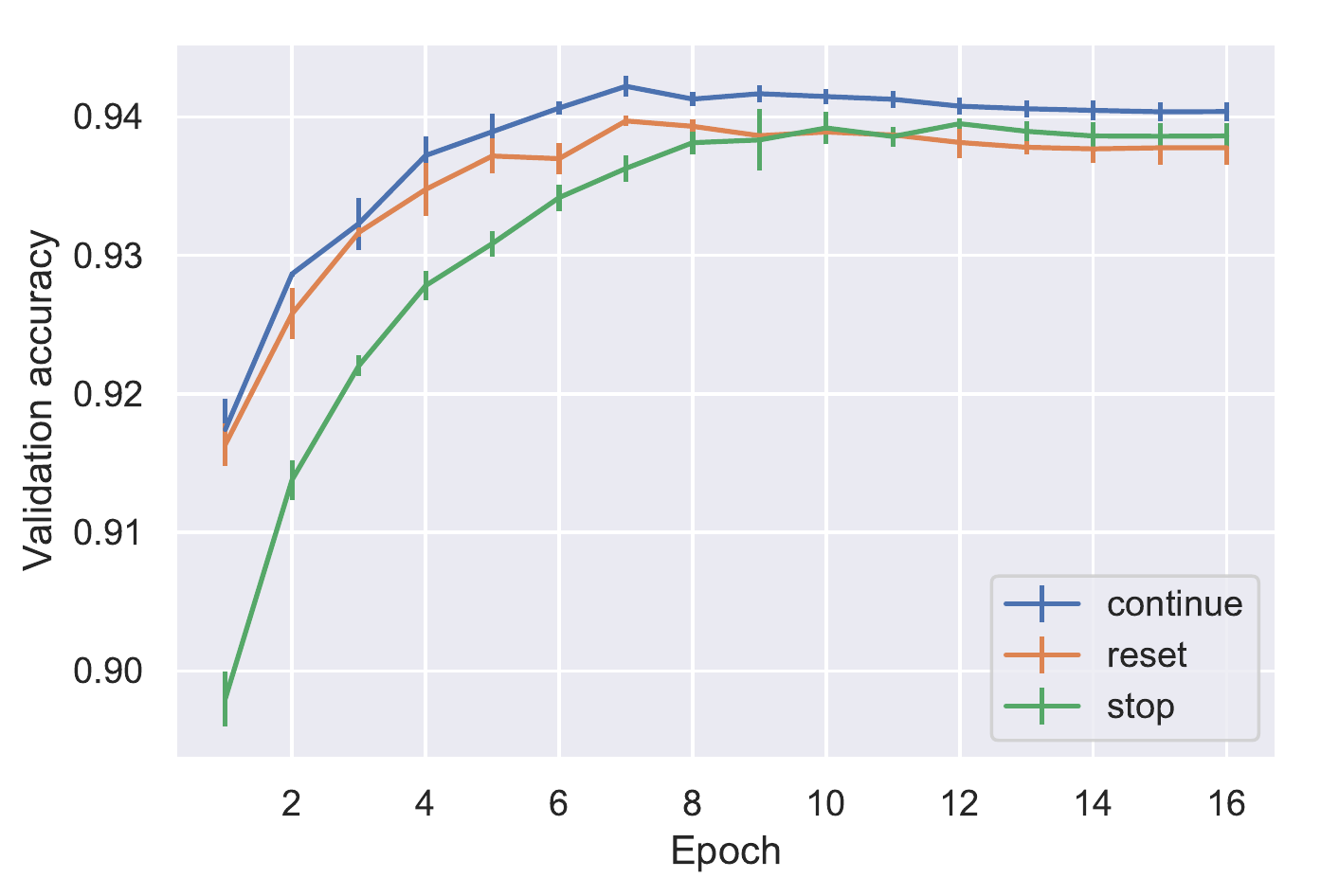}
  \includegraphics[width=0.24\textwidth]{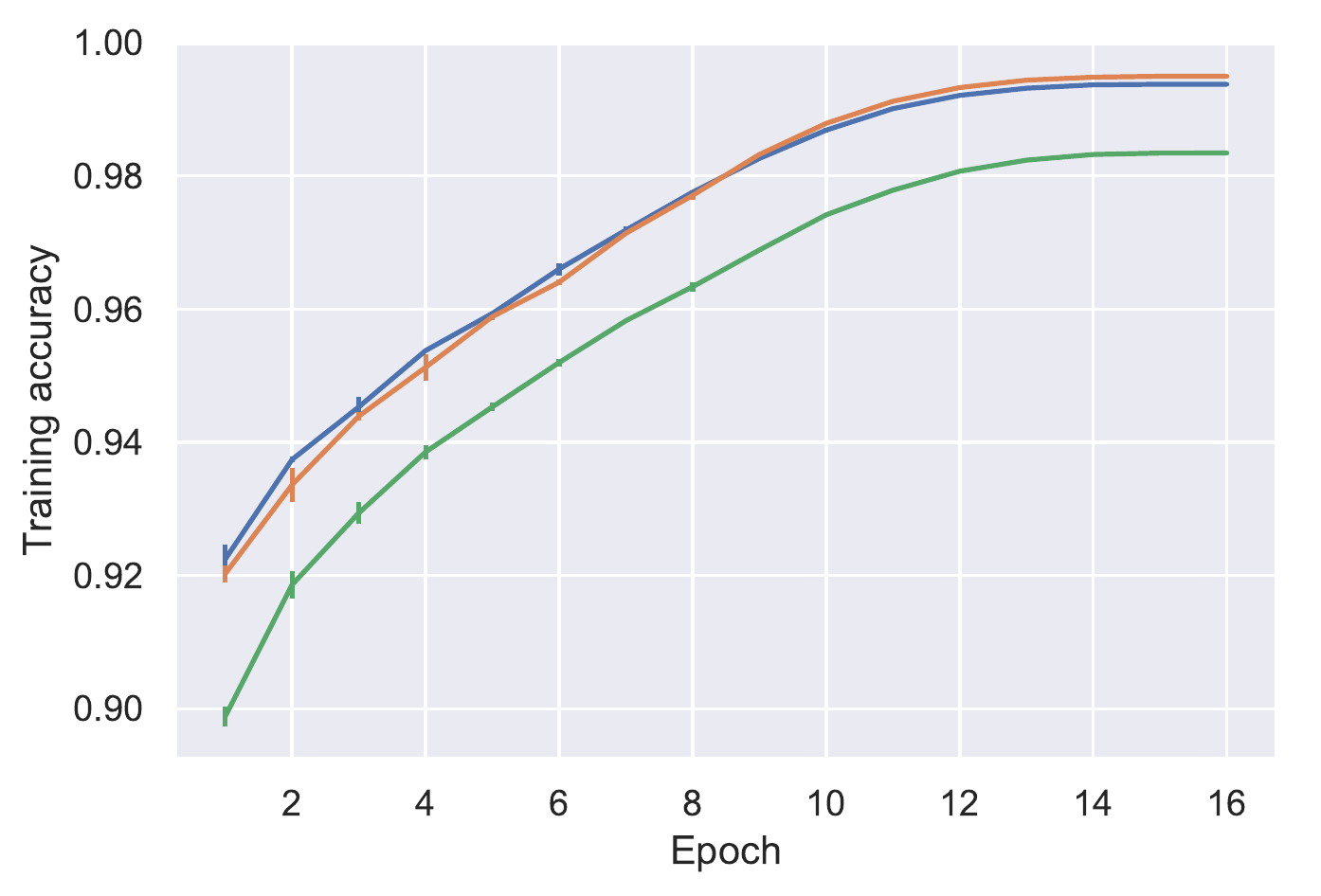}
  \includegraphics[width=0.24\textwidth]{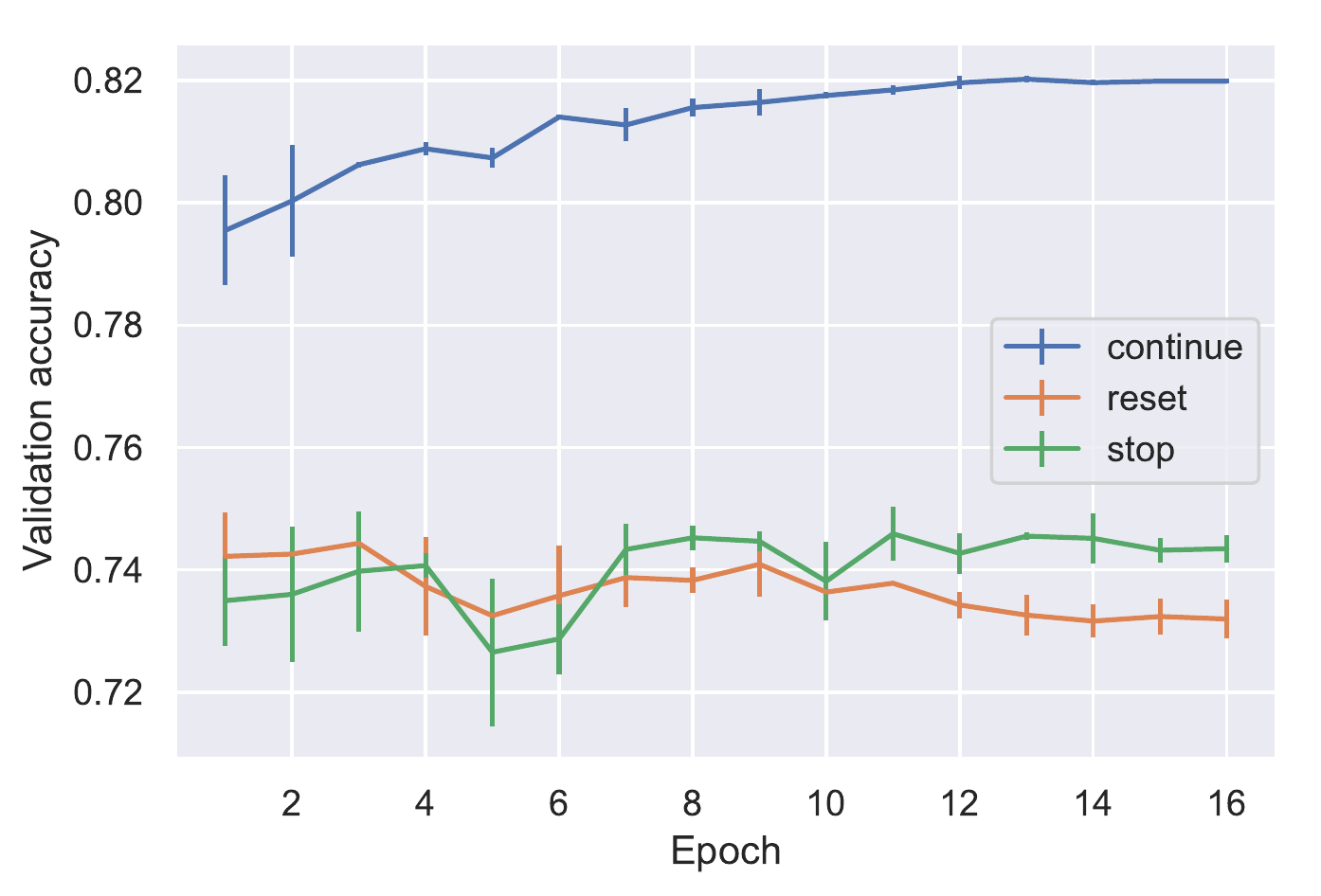}
  \includegraphics[width=0.24\textwidth]{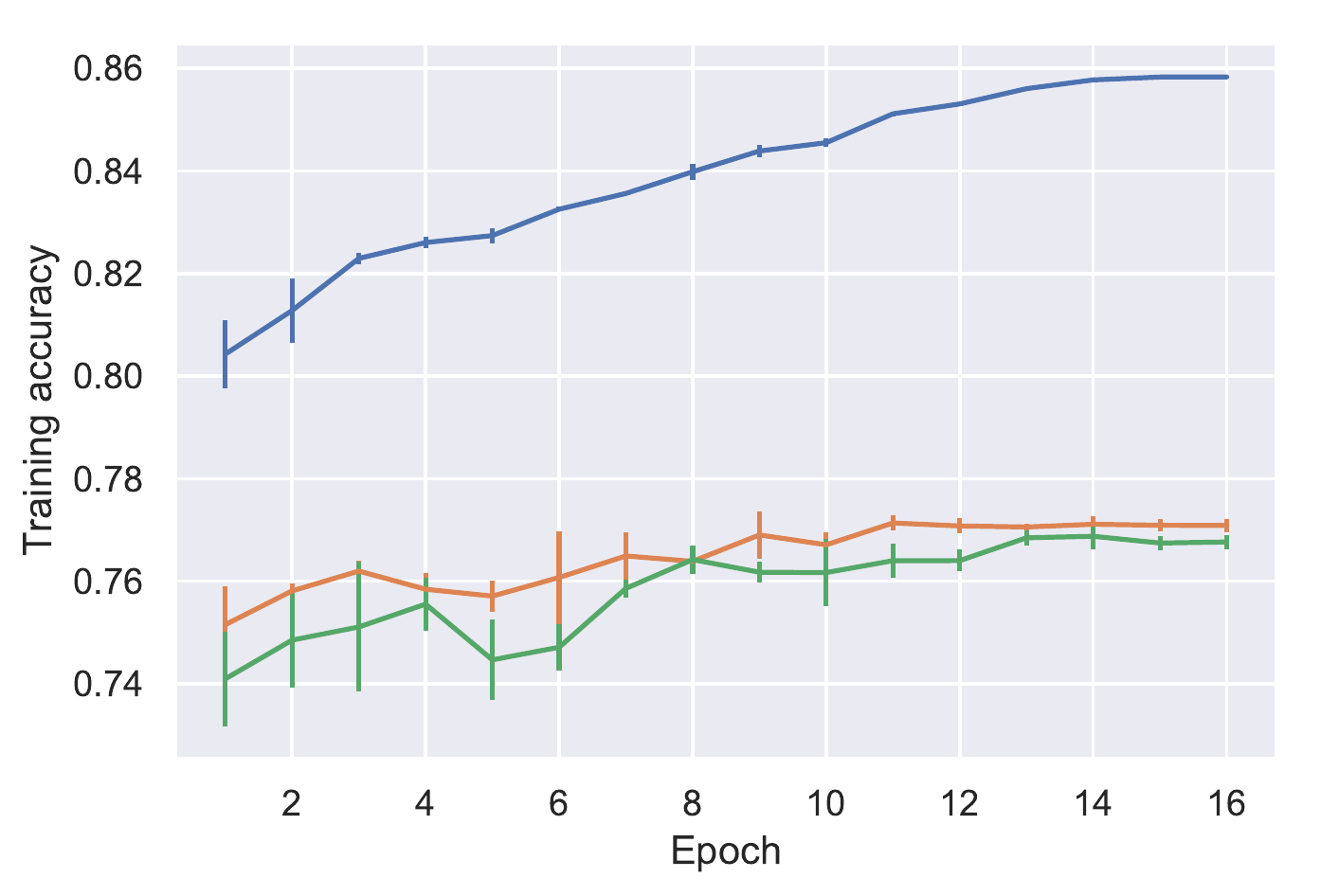}
  \caption{Validation and training accuracies for non-beam-aware training (i.e., $k = 1$) with different data collection strategies for the main (left half) and simplified (right half) models.
  \textit{continue} achieves higher accuracies.
  }
  \label{fig:non_beam_vaswani}
\end{figure*}

The gains achieved by the main model by decoding with beam search post-training are very small (from $0.02$ to $0.05$).
This suggests that training the model in a non-beam-aware fashion and then using beam search does not guarantee improvements.
The model must be learned with search to improve on these results.
For the simpler model, larger improvements are observed (from $0.42$ to $4.34$).
Despite the gains with beam search for \textit{reset} and \textit{stop}, they are not sufficient to beat the greedy model trained on its own trajectories, yielding $81.99$ for \textit{continue} with $k = 1$ versus $77.54$ for \textit{oracle} and $77.82$ for \textit{reset}, both with $k = 8$.
These results show the importance of the data collection strategy, even when the model is not trained in a beam-aware fashion.
These gains are eclipsed by beam-aware training, namely, compare Table~\ref{tab:results_nonbeam_aware} with Table~\ref{tab:results_strategies}.
See Figure~\ref{fig:non_beam_vaswani} for the evolution of the validation and training accuracies with epochs.


\begin{table}[tb]
	\centering
  \resizebox{\columnwidth}{!}{
	\begin{tabular}{lcccc}
		\toprule
	    & 1 & 2 & 4 & 8 \\
		\midrule
      oracle              & $94.10_{0.08}$ & $92.98_{0.07}$ & $91.66_{0.22}$ & $85.95_{0.79}$ \\
      reset               & $94.20_{0.11}$ & $94.34_{0.06}$ & $94.33_{0.01}$ & $94.42_{0.04}$ \\
      reset (mult.)       & $94.15_{0.07}$ & $93.98_{0.08}$ & $94.06_{0.06}$ & $94.16_{0.05}$ \\
      continue            & $94.15_{0.02}$ & $94.35_{0.05}$ & $94.37_{0.04}$ & $94.33_{0.04}$ \\
      stop                & $93.95_{0.09}$ & $94.11_{0.05}$ & $94.24_{0.07}$ & $94.25_{0.06}$ \\
    \midrule
      oracle              & $75.09_{0.17}$ & $80.67_{0.40}$ & $78.69_{1.27}$ & $47.38_{1.79}$ \\
      reset               & $75.06_{0.16}$ & $87.21_{0.14}$ & $91.24_{0.02}$ & $92.46_{0.09}$ \\
      reset (mult.)       & $75.04_{0.18}$ & $86.19_{0.12}$ & $90.76_{0.11}$ & $92.16_{0.03}$ \\
		  continue            & $82.01_{0.06}$ & $89.17_{0.08}$ & $91.80_{0.12}$ & $92.69_{0.01}$ \\
      stop                & $75.08_{0.54}$ & $87.16_{0.08}$ & $90.98_{0.13}$ & $92.18_{0.06}$ \\
    \bottomrule
  \end{tabular}
  }
	\caption{
		Development accuracies for beam-aware training with varying data collection strategies.}
	\label{tab:results_strategies}
\end{table}

\subsection{Comparing data collection strategies}
\label{ssec:comp_data_collection}

We train both models using the \textit{log loss (neighbors)}, described in Section~\ref{ssec:surrogate_losses}, and vary the data collection strategy, described in Section~\ref{ssec:data_collection}, and beam size.
Results are presented in Table~\ref{tab:results_strategies}
Contrary to Section~\ref{ssec:nonbeam_aware_training}, these models are trained to use beam search rather than it being an artifact of approximate decoding.
Beam-aware training under \emph{oracle} worsens performance with increasing beam size (due to increasing exposure bias).
During training, the model learns to pick the best neighbors for beams containing only close to optimal hypotheses, which are likely very different from the beams encountered when decoding.
The results for the simplified model are similar---with increasing beam size, performance first improves but then degrades.
For the main model, we observe modest but consistent improvements with larger beam sizes across all data collection strategies except \textit{oracle}.
By comparing the results with those in the first row of Table~\ref{tab:results_nonbeam_aware}, we see that we improve on the model trained with maximum likelihood and decoded with beam search.


%
\begin{table}[tb]
	\centering
  \resizebox{\columnwidth}{!}{
	\begin{tabular}{lcccc}
		\toprule
	    & 1 & 2 & 4 & 8 \\
		\midrule
        percep. (first)         & $92.81_{0.06}$ & $93.22_{0.04}$ & $93.44_{0.02}$ & $93.52_{0.06}$ \\
        percep. (last)          & $92.84_{0.11}$ & $93.57_{0.06}$ & $93.86_{0.09}$ & $93.77_{0.04}$ \\
        m. (last)               & $94.10_{0.07}$ & $94.29_{0.07}$ & $94.27_{0.03}$ & $94.43_{0.04}$ \\
        cost-s. m. (last)       & $93.98_{0.03}$ & $94.32_{0.10}$ & $94.37_{0.03}$ & $94.33_{0.13}$ \\
        log loss (beam)         & $92.29_{0.07}$ & $92.09_{0.11}$ & $94.24_{0.08}$ & $94.32_{0.02}$ \\
        log loss (neig.)        & $94.22_{0.00}$ & $94.29_{0.03}$ & $94.27_{0.06}$ & $94.38_{0.01}$ \\
        \midrule
        percep. (first)         & $77.62_{0.14}$ & $86.32_{0.05}$ & $89.83_{0.05}$ & $91.00_{0.07}$ \\
        percep. (last)          & $77.67_{0.07}$ & $87.62_{0.03}$ & $90.82_{0.16}$ & $91.98_{0.11}$ \\
        m. (last)               & $81.75_{0.04}$ & $88.80_{0.02}$ & $91.91_{0.05}$ & $92.81_{0.05}$ \\
        cost-s. m. (last)       & $81.76_{0.05}$ & $88.92_{0.06}$ & $91.81_{0.03}$ & $92.81_{0.03}$ \\
        log loss (beam)         & $77.50_{0.07}$ & $88.25_{0.08}$ & $91.46_{0.06}$ & $92.56_{0.11}$ \\
        log loss (neig.)        & $81.94_{0.02}$ & $89.01_{0.10}$ & $91.75_{0.03}$ & $92.60_{0.03}$ \\
	    \bottomrule
  \end{tabular}
  }
	\caption{Development accuracies for the loss functions in Section~\ref{ssec:surrogate_losses}.}
	\label{tab:results_losses}
\end{table}

\begin{figure*}[tbp]
  \centering
  \includegraphics[width=0.24\textwidth]{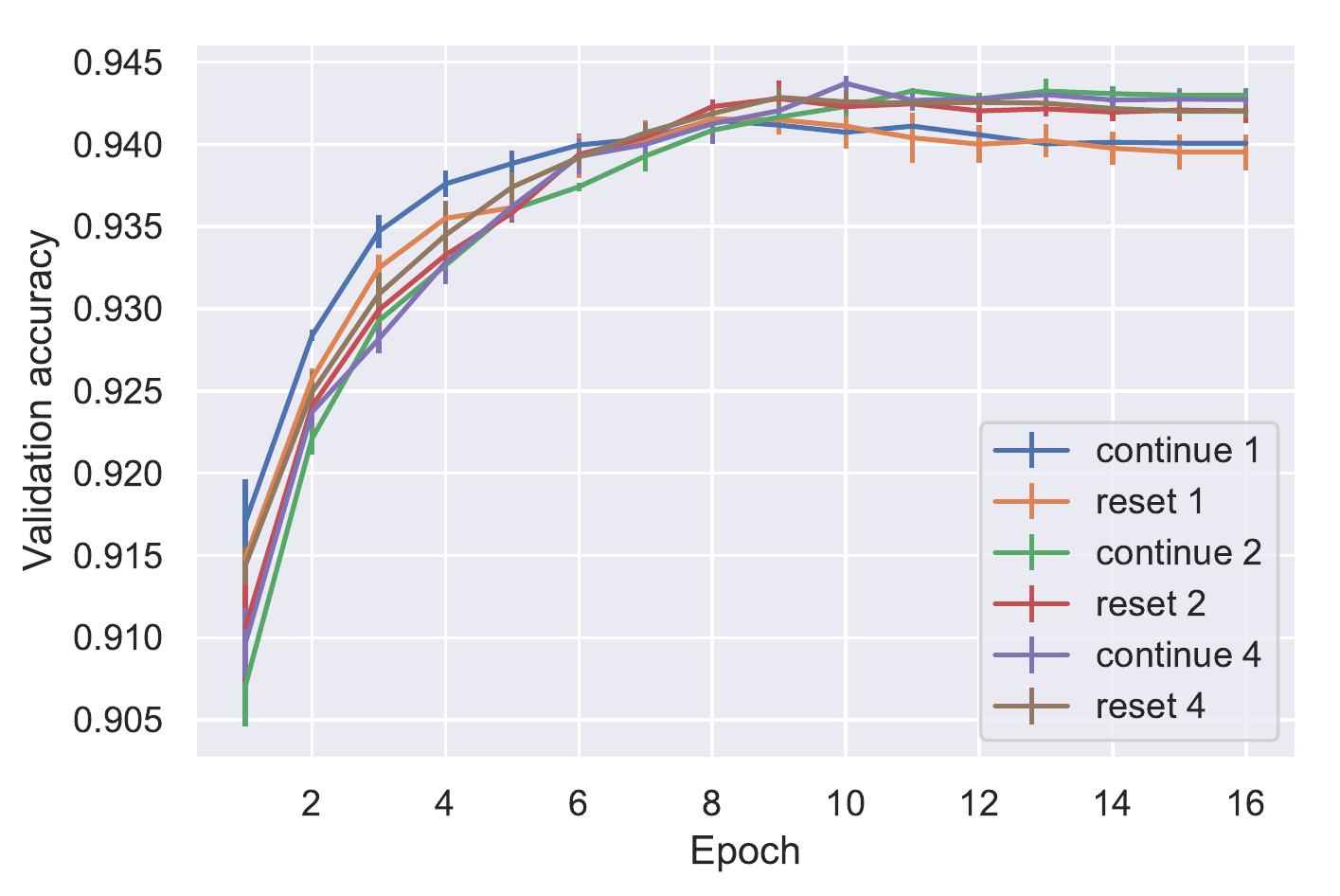}
  \includegraphics[width=0.24\textwidth]{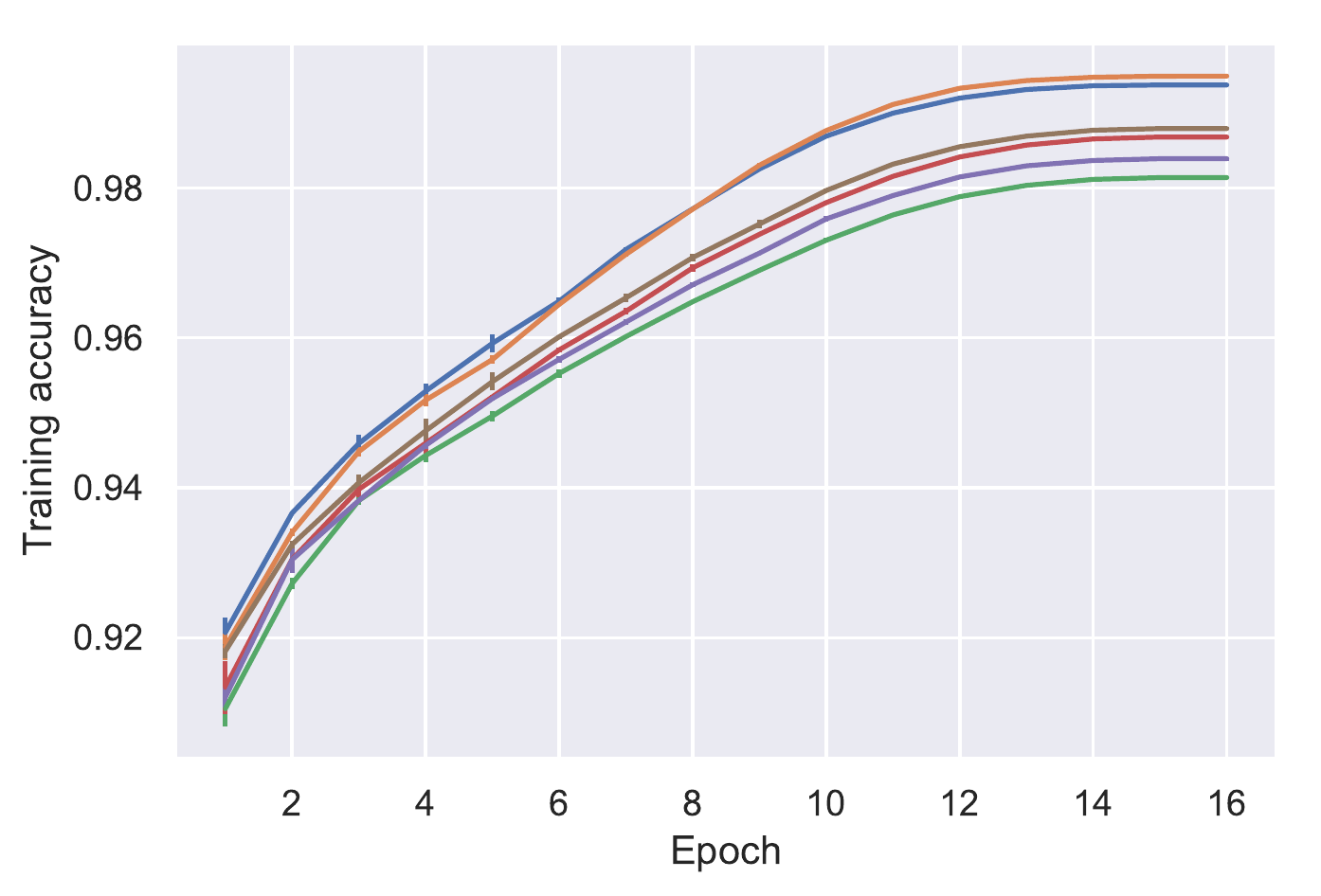}
  \includegraphics[width=0.24\textwidth]{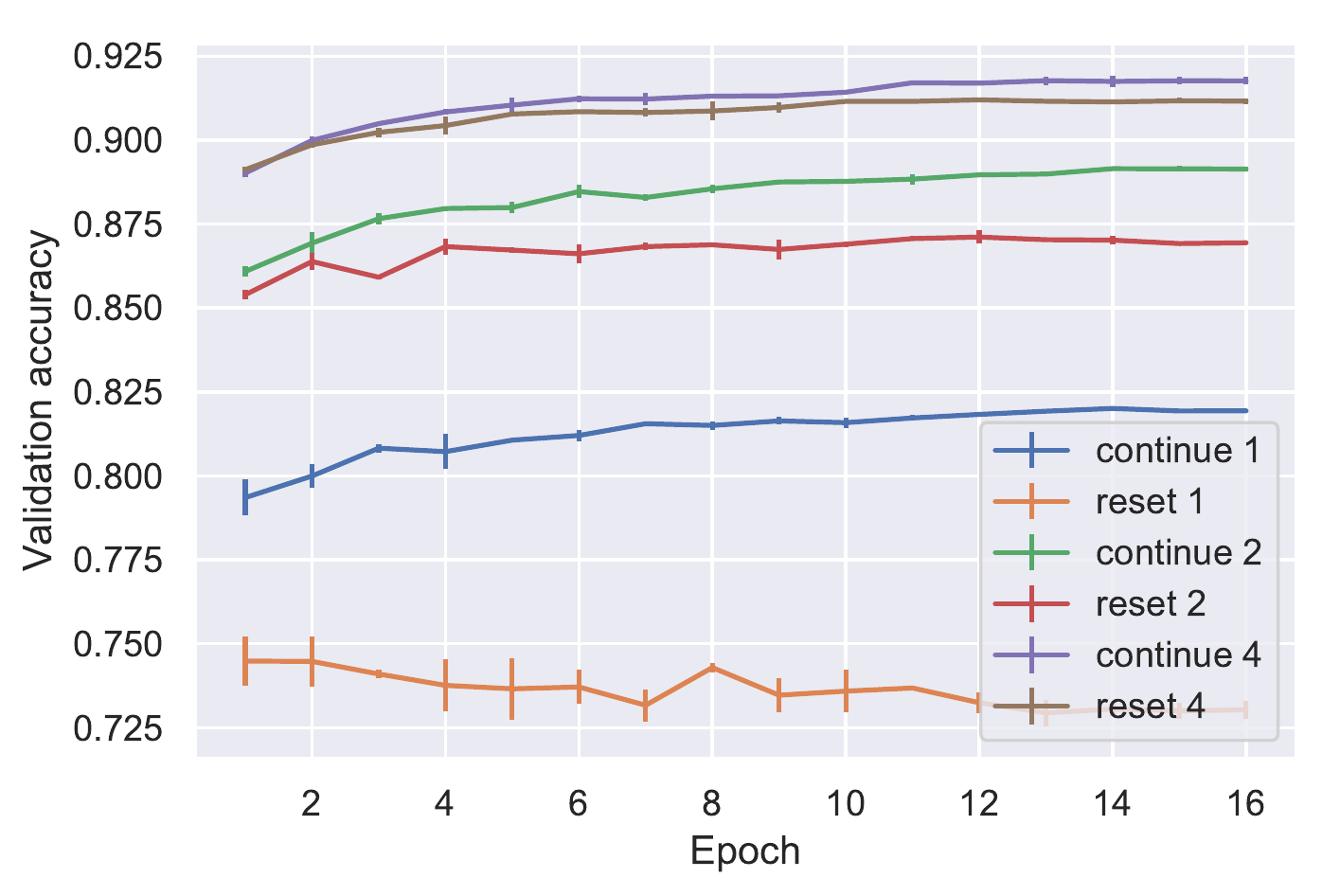}
  \includegraphics[width=0.24\textwidth]{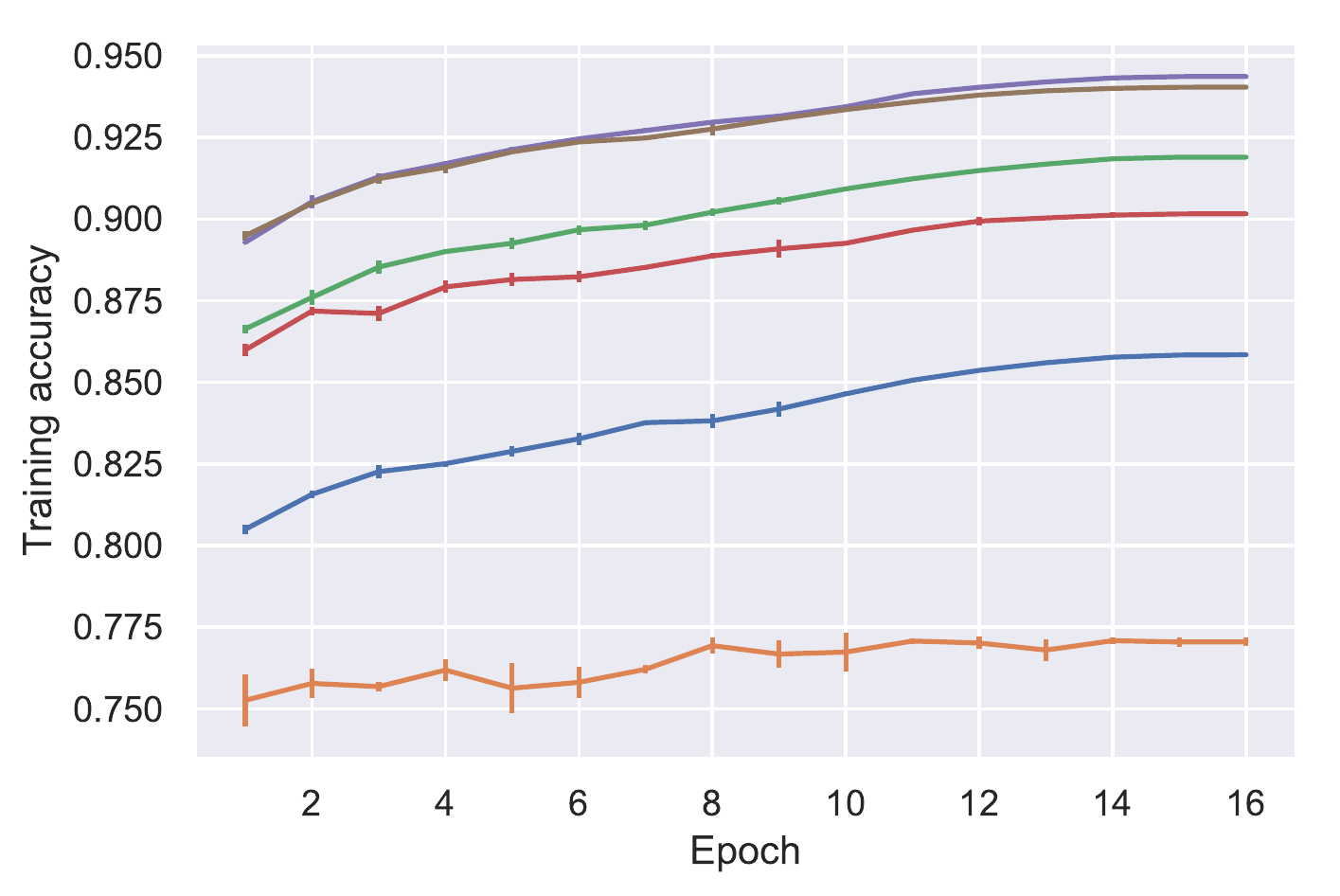}
  \caption{Validation and training accuracies for beam-aware training with different data collection strategies and beam sizes for the main (left half) and simplified (right half) models.
  Larger beam sizes achieve higher performances while overfitting less, and are crucial for the simplified model to achieve higher training and validation accuracies.
  For smaller beams \textit{continue} performs better than \textit{reset}.
  All models can be trained stably from scratch.
  Three runs were aggregated by showing the mean and the standard deviation for each epoch.
  }
  \label{fig:beam_vaswani}
\end{figure*}

The data collection strategy has a larger impact on performance for the simplified model.
\textit{continue} achieves the best performance.
Compare these performances with those for the simplified model in Table~\ref{tab:results_nonbeam_aware}.
For larger beams, the improvements achieved by beam-aware training are much larger than those achieved by non-beam-aware ones.
For example, $92.69$ versus $82.41$ for \textit{continue} with $k = 8$, where in the first case it is trained in a beam-aware manner ($k=8$ for both training and decoding), while in the second case, beam search is used only during decoding ($k = 1$ during training but $k = 8$ during decoding).
This shows the importance of training with beam search and exposing the model to its mistakes.
Without beam-aware training, the model is unable to learn to use the beam effectively.
Check Figure~\ref{fig:beam_vaswani} for the evolution of the training and validation accuracies with training epoch for beam-aware training.

\subsection{Comparing surrogate losses}
\label{ssec:comp_surrogate_losses}

We train both models with \textit{continue} and vary the surrogate loss and beam size.
Results are presented in Table~\ref{ssec:surrogate_losses}.
Perceptron losses (e.g., \textit{perceptron (first)} and \textit{perceptron (last)}) performed worse than their margin-based counterparts (e.g., \textit{margin (last)} and \textit{cost-sensitive margin (last)}).
\textit{log loss (beam)} yields poor performances for small beam sizes (e..g, $k=1$ and $k = 2$).
This is expected due to small contrastive sets (i.e., at most $k+1$ elements are used in \textit{log loss (beam)}).
For larger beams, the results are comparable with \textit{log loss (neighbors)}.

\subsection{Additional design choices}
\label{sec:discussion}

\paragraph{Score accumulation}

The scoring function was introduced as a sum of prefix terms.
A natural alternative is to produce the score for a neighbor without adding it to a running sum, i.e., $s(y_{1:j}, \theta) = \tilde s(y_{1:j}, \theta)$ rather than $s(y_{1:j}, \theta) = \sum_{i = 1} ^j \tilde s(y_{1:i}, \theta)$.
Surprisingly, score accumulation performs uniformly better across all configurations.
For the main model, beam-aware training degraded performance with increasing beam size.
For the simplified model, beam-aware training improved on the results in Table~\ref{tab:results_nonbeam_aware}, but gains were smaller than those with score accumulation.
We observed that the LM LSTM failed to keep track of differences earlier in the supertag sequence, leading to similar scores over their neighbors.
Accumulating the scores is a simple memory mechanism that does not require the LM LSTM to learn to propagate long-range information.
This performance gap may not exist for models that access information more directly (e.g., transformers~\citep{vaswani2017attention} and other attention-based models~\citep{bahdanau2014neural}).
See the appendix for Table~\ref{tab:results_score_accumulation} which compares configurations with and without score accumulation.
Performance differences range from $1$ to $5$ absolute percentage points.

\paragraph{Update on all beams}
The meta-algorithm of~\citet{negrinho2018learning} suggests inducing losses on every visited beam as there is always a correct action captured by appropriately scoring the neighbors.
This leads to updating the parameters on every beam.
By contrast, other beam-aware work updates only on beams where the transition leads to increased cost (e.g., \citet{daume2005learning} and \citet{andor2016globally}).
We observe that always updating leads to improved performance, similar to the results in Table~\ref{tab:results_losses} for perceptron losses.
We therefore recommend inducing losses on every visited beam.
See the appendix for Table~\ref{tab:results_only_on_cost_increase}, which compares configurations trained with and without updating on every beam.

\section{Related work}
\label{sec:related_work}

Related work uses either imitation learning (often called learning to search when applied to structured prediction) or beam-aware training.
Learning to search~\cite{daume2009search-based, chang2015learning, goldberg2012dynamic, bengio2015scheduled, negrinho2018learning} is a popular approach for structured prediction.
This literature is closely related to imitation learning~\cite{ross2010efficient, ross2011reduction, ross2014reinforcement}.
\citet{ross2011reduction} addresses exposure bias by collecting data with the learned policy at training time.
\citet{collins_incremental_2004} proposes a structured perceptron variant that trains with beam search, updating the model parameters when the correct hypothesis falls out of the beam.
\citet{huang2012structured} introduces a theoretical framework to analyze the convergence of early update.
\citet{zhang2008tale} develops a beam-aware algorithm for dependency parsing that uses early update and dynamic oracles.
\citet{goldberg2012dynamic, goldberg2013training} introduce dynamic oracles for dependency parsing.
\citet{ballesteros2016training} observes that exposing the model to mistakes during training improves a dependency parser.
\citet{bengio2015scheduled} makes a similar observation and present results on image captioning, constituency parsing, and speech recognition.
Beam-aware training has also been used for speech recognition~\cite{collobert2019fully, baskar2019promising}.
\citet{andor2016globally} proposes an early update style algorithm for learning models with a beam, but use a log loss rather than a perceptron loss as in \citet{collins_incremental_2004}.
Parameters are updated when the golden hypothesis falls out of the beam or when the model terminates with the golden hypothesis in the beam.
\citet{wiseman2016sequence} use a similar algorithm to \citet{andor2016globally} but they use a margin-based loss and reset to a beam with the golden hypothesis when it falls out of the beam.
\citet{edunov2017classical} use beam search to find a contrastive set to define sequence-level losses.
\citet{goyal2017continuous, goyal2019empirical} propose a beam-aware training algorithm that relies on a continuous approximation of beam search.
\citet{negrinho2018learning} introduces a meta-algorithm that instantiates beam-aware algorithms based on choices for beam size, surrogate loss function, and data collection strategy.
They propose a DAgger-like algorithm for beam search.

\section{Conclusions}
\label{sec:conclusions}

Maximum likelihood training of locally normalized models with beam search decoding is the default approach for structured prediction.
Unfortunately, it suffers from exposure bias and does not learn to use the beam effectively.
Beam-aware training promises to address some of these issues, but is not yet widely used due to being poorly understood.
In this work, we explored instantiations of the meta-algorithm of \citet{negrinho2018learning} to understand how design choices affect performance.
We show that beam-aware training is most useful when substantial uncertainty must be managed during prediction.
We make recommendations for instantiating beam-aware algorithms based on the meta-algorithm, such as inducing losses at every beam, using log losses (rather than perceptron-style ones), and preferring the \textit{continue} data collection strategy (or \textit{reset} if necessary).
We hope that this work provides evidence that beam-aware training can greatly impact performance and be trained stably, leading to their wider adoption.

\section*{Acknowledgements}

We gratefully acknowledge support from 3M | M*Modal.
This work used the Bridges system, which is supported by NSF award number ACI-1445606, at the Pittsburgh Supercomputing Center (PSC).

\bibliographystyle{acl_natbib}
\bibliography{emnlp2020}

\appendix

\clearpage

\onecolumn

\section{Additional results}


\begin{table}[H]
	\centering
	\begin{tabular}{lcccc}
		\toprule
    & 1 & 2 & 4 & 8 \\
		\midrule
    log loss (neighbors); no acc.            & $94.25_{0.02}$ & $93.61_{0.07}$ & $92.77_{0.11}$ & $92.08_{0.08}$ \\
    log loss (beam); no acc.                 & $92.31_{0.08}$ & $86.89_{0.20}$ & $87.58_{0.27}$ & $89.51_{0.28}$ \\
    cost-sensitive margin (last); no acc.    & $93.94_{0.11}$ & $93.31_{0.06}$ & $92.30_{0.05}$ & $90.61_{0.14}$ \\
    \midrule
    log loss (neighbors); acc.               & $94.23_{0.03}$ & $94.30_{0.06}$ & $94.31_{0.05}$ & $94.37_{0.09}$ \\
    log loss (beam); acc.                    & $92.33_{0.08}$ & $92.16_{0.50}$ & $94.26_{0.05}$ & $94.37_{0.05}$ \\
    cost-sensitive margin (last); acc.       & $94.07_{0.07}$ & $94.26_{0.07}$ & $94.28_{0.04}$ & $94.26_{0.04}$ \\
    \midrule
    \midrule
    log loss (neighbors); no acc.            & $81.95_{0.03}$ & $88.01_{0.06}$ & $89.48_{0.08}$ & $88.77_{0.09}$ \\
    log loss (beam); no acc.                 & $77.70_{0.04}$ & $86.76_{0.06}$ & $88.58_{0.09}$ & $88.12_{0.06}$ \\
    cost-sensitive margin (last); no acc.    & $81.74_{0.05}$ & $87.35_{0.11}$ & $89.18_{0.12}$ & $88.89_{0.17}$ \\
    \midrule
    log loss (neighbors); acc.               & $82.00_{0.05}$ & $89.14_{0.08}$ & $91.68_{0.02}$ & $92.60_{0.04}$ \\
    log loss (beam); acc.                    & $77.61_{0.26}$ & $88.18_{0.05}$ & $91.44_{0.06}$ & $92.58_{0.03}$ \\
    cost-sensitive margin (last); acc.       & $81.73_{0.01}$ & $88.83_{0.02}$ & $91.85_{0.03}$ & $92.87_{0.02}$ \\
  \bottomrule
	\end{tabular}
	\caption{Trained with and without score accumulation.}
	\label{tab:results_score_accumulation}
\end{table}


\begin{table}[H]
	\centering
	\begin{tabular}{lcccc}
		\toprule
    & 1 & 2 & 4 & 8 \\
		\midrule
    log loss (neighbors); always                & $94.27_{0.10}$ & $94.37_{0.02}$ & $94.39_{0.06}$ & $94.33_{0.07}$ \\
    log loss (beam); always                     & $92.26_{0.08}$ & $92.23_{0.66}$ & $94.33_{0.05}$ & $94.30_{0.04}$ \\
    \midrule
    log loss (neighbors); on-cost increase      & $92.06_{0.78}$ & $93.65_{0.07}$ & $93.67_{0.16}$ & $92.22_{0.32}$ \\
    log loss (beam); on-cost increase           & $92.23_{0.11}$ & $93.00_{0.33}$ & $93.10_{0.57}$ & $91.98_{0.32}$ \\
    \midrule
    \midrule
    log loss (neighbors); always                & $82.01_{0.04}$ & $89.11_{0.05}$ & $91.75_{0.01}$ & $92.62_{0.06}$ \\
    log loss (beam); always                     & $77.52_{0.17}$ & $88.19_{0.05}$ & $91.43_{0.03}$ & $92.49_{0.04}$ \\
    \midrule
    log loss (neighbors); on-cost increase      & $79.35_{0.07}$ & $88.53_{0.08}$ & $91.44_{0.03}$ & $92.45_{0.06}$ \\
    log loss (beam); on-cost increase           & $77.57_{0.06}$ & $86.69_{0.07}$ & $90.47_{0.03}$ & $91.72_{0.05}$ \\
  \bottomrule
	\end{tabular}
  \caption{Trained with updating on cost-increase versus always updating.
  }
	\label{tab:results_only_on_cost_increase}
\end{table}
%


\begin{table}[H]
	\centering
	\begin{tabular}{lcccc}
		\toprule
	    & 1 & 2 & 4 & 8 \\
		\midrule
      oracle              & $94.09_{0.09}$ & $92.92_{0.11}$ & $91.29_{0.27}$ & $86.47_{0.62}$ \\
      reset               & $94.06_{0.07}$ & $94.15_{0.07}$ & $94.16_{0.02}$ & $94.14_{0.07}$ \\
      reset (mult.)       & $93.97_{0.03}$ & $93.87_{0.04}$ & $93.95_{0.01}$ & $94.05_{0.09}$ \\
      continue            & $94.16_{0.03}$ & $94.23_{0.09}$ & $94.23_{0.04}$ & $94.29_{0.11}$ \\
      stop                & $93.88_{0.02}$ & $94.02_{0.04}$ & $94.09_{0.06}$ & $94.18_{0.07}$ \\
    \midrule
      oracle              & $74.55_{0.21}$ & $81.54_{0.43}$ & $77.18_{0.74}$ & $49.67_{3.23}$ \\
      reset               & $75.38_{0.44}$ & $87.00_{0.14}$ & $90.98_{0.05}$ & $92.31_{0.18}$ \\
      reset (mult.)       & $74.85_{0.46}$ & $86.19_{0.17}$ & $90.45_{0.11}$ & $91.78_{0.07}$ \\
		  continue            & $81.55_{0.01}$ & $88.86_{0.02}$ & $91.53_{0.19}$ & $92.36_{0.06}$ \\
      stop                & $74.61_{0.35}$ & $86.66_{0.10}$ & $90.78_{0.12}$ & $91.97_{0.04}$ \\
    \bottomrule
	\end{tabular}
	\caption{
    Trained for double the number of epochs as those in Table~\ref{tab:results_strategies}.
    }
	\label{tab:results_strategies_train32epochs}
\end{table}

\end{document}